# SIRAG: Towards Stable and Interpretable RAG with A Process-Supervised Multi-Agent Framework

Junlin Wang†
Heyuan Tobacco Monopoly Administration
Xinyuan Road, Yuancheng District
Heyuan, China
953620519@qq.com

Zehao Wu†
School of Automation Science and Engineering
South China University of Technology
Guangzhou, China
auzhwu@mail.scut.edu.cn

Shaowei Lu
Heyuan Tobacco Monopoly Administration
Xinyuan Road, Yuancheng District
Heyuan, China
416149366@qq.com

Yanlan Li
Heyuan Tobacco Monopoly Administration
Xinyuan Road, Yuancheng District
Heyuan, China
191732947@qq.com

Xinghao Huang*
Heyuan Tobacco Monopoly Administration
Xinyuan Road, Yuancheng District
Heyuan, China
*Corresponding author: ouxiangpoul@126.com
†: These authors contributed equally to this work

*Abstract*—Retrieval-Augmented Generation (RAG) enables large language models (LLMs) to access external knowledge sources, but the effectiveness of RAG relies on the coordination between the retriever and the generator. Since these components are developed independently, their interaction is often suboptimal: the retriever may return irrelevant or redundant documents, while the generator may fail to fully leverage retrieved evidence. In this work, we propose a process-supervised multi-agent framework to bridge the gap between retriever and generator. The framework introduces two lightweight agents: a Decision Maker, which determines when to continue retrieval or stop for answer generation, and a Knowledge Selector, which filters retrieved documents to retain only the most useful evidence. To provide fine-grained supervision, we employ an LLM-as-a-Judge that evaluates each intermediate action with process-level rewards, ensuring more accurate credit assignment than relying solely on final answer correctness. We further adopt a tree-structured rollout strategy to explore diverse reasoning paths, and train both agents with Proximal Policy Optimization (PPO) in an end-to-end manner. Experiments on single-hop and multi-hop question answering benchmarks show that our approach achieves higher accuracy, more stable convergence, and produces more interpretable reasoning trajectories compared with standard RAG baselines. Importantly, the proposed framework is modular and plug-and-play, requiring no modification to the retriever or generator, making it practical for real-world RAG applications.

*Keywords—Retrieval-Augmented Generation (RAG), Multi-Agent Cooperation, Proximal Policy Optimization (PPO)*

## I. INTRODUCTION

Large language models (LLMs) have demonstrated remarkable capabilities in knowledge-intensive tasks such as open-domain question answering and reasoning. However, they remain limited by their static parametric knowledge and are prone to hallucinations.[1]-[3] To mitigate these issues, **Retrieval-Augmented Generation (RAG)** has become a widely adopted paradigm [4], where an external retriever fetches potentially relevant documents to ground the generation processs[5]. Despite its success, the effectiveness of RAG still hinges on the tight coordination between the retriever and the generator. Since these components are typically developed independently, they often suffer from semantic and functional misalignment: the retriever may provide documents that are irrelevant or redundant, while the generator may fail to formulate effective queries or fully leverage the retrieved evidence.

Existing approaches attempt to bridge this gap through retriever fine-tuning, generator adaptation, or introducing intermediate modules such as rerankers or query rewriters[6]-[11]. While these methods offer partial improvements, they also face notable limitations: retriever fine-tuning requires curated data and cannot be easily applied to commercial search engines[12][13], generator fine-tuning is computationally expensive and risks degrading pretrained capabilities[14], and task-specific intermediate modules usually optimize only a single stage, leading to suboptimal coordination across the entire pipeline[6][14]. Thus, there is a strong need for a **lightweight yet general mechanism** that can flexibly coordinate retriever and generator behaviors without retraining them.

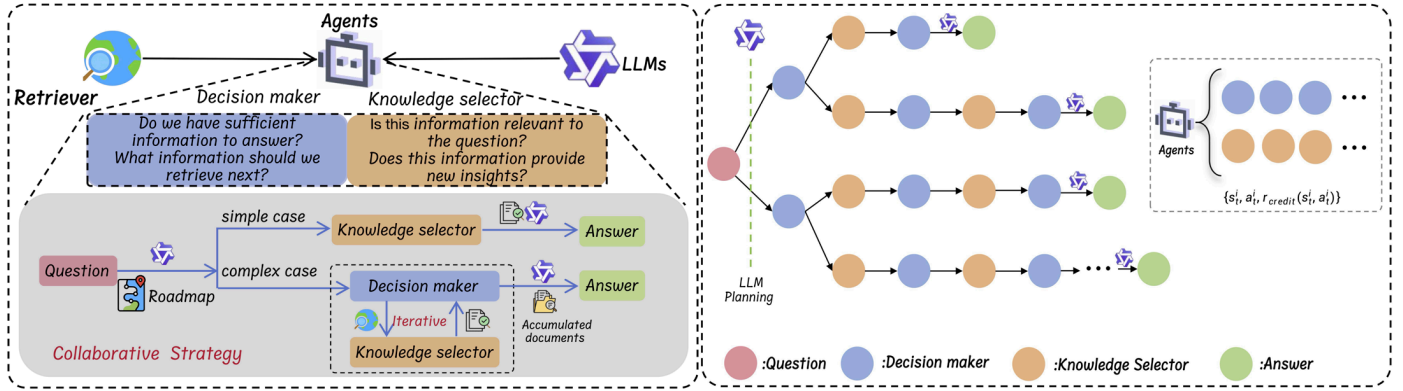

Figure 1: Overall framework of SIRAG. Left: The proposed lightweight multi-agent system with collaborative strategies. Right: The end-to-end optimization pipeline for our multi-agent system.

To address this challenge, we propose a **process-supervised multi-agent framework** that explicitly models and optimizes the decision process connecting retrieval and generation. Our framework introduces two lightweight agents: a **Decision Maker**, which decides whether to continue retrieval, reformulate queries, or hand over to the generator, and a **Knowledge Selector**, which filters retrieved documents to ensure that only the most relevant and useful evidence is passed to the generator. These agents interact cooperatively to form reasoning trajectories that bridge the gap between retriever and generator.

A key novelty of our approach is the use of **LLM-as-Judge** to provide **process-level supervision**. Instead of relying solely on the correctness of the final generated answer as a sparse reward signal, we employ a strong LLM to evaluate each intermediate action, assigning fine-grained credit to decisions such as query formulation and evidence selection. This not only mitigates the credit assignment problem but also improves training stability and interpretability. To further enhance learning efficiency, we adopt a **tree-structured rollout strategy** to explore multiple reasoning paths, and optimize agent policies end-to-end with **Proximal Policy Optimization (PPO)**.

We validate our method on multiple single-hop and multi-hop question answering benchmarks. Experimental results demonstrate that our process-supervised proxy framework achieves superior answer accuracy compared to strong baselines, while also yielding more interpretable reasoning trajectories. Our key contributions are as follows:

- We propose a multi-agent framework with Decision Maker and Knowledge Selector agents to coordinate retriever–generator interactions.
- We introduce LLM-as-a-Judge process supervision, providing fine-grained rewards for intermediate actions to improve stability and interpretability.
- We develop a tree-structured rollout with PPO optimization to enable effective end-to-end training of cooperative agents.
- Experiments on diverse QA benchmarks show improved accuracy and reasoning quality, while maintaining plug-and-play modularity.

## II. METHODS

In this section, we provide a detailed explanation of the implementation of SIRAG, with the overall architecture illustrated in Figure 1.

### A. Problem Formulation

We consider the task of retrieval-augmented generation(RAG), where the goal is to answer a question $q$ by leveraging both an external retriever and a generator. Let $\mathcal{R}(q)$ denote the retriever returning a set of documents, and $\mathcal{G}(\cdot)$ denote the generator producing the final answer. The key challenge is to ensure that the retrieval process and the generator are well aligned so that the retrieved documents are useful for the generation step.

Instead of directly optimizing the retriever or the generator, we introduce a multi-agent proxy system to control the intermediate decision-making process. Formally, we design two cooperative agents:

- **Decision Maker (DM)**: decides whether to issue a new query, terminate retrieval, or hand over to the generator.
- **Knowledge Selector (KS)**: filters the retrieved documents to select the most relevant subset for the generator.

Each agent operate in a partially observable environment and interacts sequentially, forming an action trajectory $\tau = \{(s_t, a_t)\}_{t=1}^T$. The objective is to learn optimal policies $\pi_\theta^{DM}$ and $\pi_\theta^{KS}$ that maximize expected task performance.

### B. Multi-agent interaction

At each step, the **Decision Maker** observes the current question, accumulated evidence, and reasoning state $s_t^{DM}$, and chooses one of the following actions:

$a_t^{DM} \in \{Retrieval(new\ query), Stop\&Generation\}$

If a retrieval action is chosen, a query is sent to the retriever, and the retrieved documents are passed to the **Knowledge**

**Selector**. The **Knowledge Selector** then observes $(q, \mathcal{R}(q), context)$ and selects a subset of documents:
$$a_t^{KS} = Select(d_1, d_2, ..., d_n) \subseteq \mathcal{R}(q)$$
These filtered documents are appended to the evidence pool. The process repeats until the Decision Maker outputs **Stop & Generate**, in which case all accumulated evidence is passed to the generator $\mathcal{G}$ for answer synthesis.

*C. LLM-as-a-Judge: Process-Level Reward Modeling*

To collect diverse trajectories for training, we adopt a **tree-structured rollout strategy**. For each input question, the Decision Maker is forced to explore multiple reasoning strategies at the top level (e.g., retrieving vs. stopping early), while deeper levels are expanded stochastically. This results in a decision tree where each path corresponds to a possible reasoning trajectory:

$$\tau = \{(s_t, a_t)\}_{t=1}^T, R(\tau) \in \{0,1\} \quad (1)$$

Where $R(\tau)$ is the system-level reward (final answer correctness). This rollout ensures that both simple and complex strategies are evaluated, providing a richer training signal than single-path exploration.

System-level rewards alone are sparse and insufficient for credit assignment. To address this, we propose using a strong LLM (e.g., GPT-4, Qwen2-72B) as a **process supervisor** to evaluate the quality of each intermediate action. For each node $(s_t, a_t)$ in the rollout tree, the LLM judge provides a score:

$$r_{process}(s_t, a_t) \in [0,1] \quad (2)$$

reflecting whether the action is reasonable, informative, and consistent with the question intent.

The final credit for each action combines system- and process-level signals:

$$r_{credit}(s_t, a_t) = \alpha \cdot R(\tau) + \beta \cdot r_{process}(s_t, a_t) \quad (3)$$

Where $\alpha, \beta$ are trade-off coefficients.

*D. Policy Optimization with PPO*

We train both agents with **Proximal Policy Optimization (PPO)**, a widely used reinforcement learning algorithm.

Having obtained the credit rewards that reflect each agent's contribution, we develop an optimization framework to guide end-to-end training across all agents. The key idea is to use these credit signals for optimizing the collaborative behavior of the entire system. The optimization objective for our multi-agent system can be formulated as maximizing the expected credit rewards:

$$\mathcal{J}(\theta) = \mathbb{E}_{\tau \in \pi_\theta}\left[\sum_{i \in \mathcal{N}} \sum_t r_{credit}(s_t^i, a_t^i)\right] \quad (4)$$

Since each agent's action is a sequence of tokens, we decompose this optimization using Proximal Policy Optimization (PPO) [16][17][18] as follows:

$$\mathcal{L}_{SIRAG} = \sum_{i \in \mathcal{N}} \mathcal{L}_{PPO}^i(\theta, \varphi) \quad (5)$$

Specifically, for each agent $i$, we define:

$$\mathcal{L}_{CLIP}^i(\theta) = \mathbb{E}_{\tau \in \pi_\theta}\left[\sum_t \sum_m \min(r_{t,m}^i(\theta)\hat{A}_{t,m}^i, clip(r_{t,m}^i(\theta), 1-\varepsilon, 1+\varepsilon)\hat{A}_{t,m}^i)\right] \quad (6)$$

Where $r_{t,m}^i(\theta) = \frac{\pi_\theta(a_{t,m}^i|s_{t,m}^i)}{\pi_{\theta_{old}}(a_{t,m}^i|s_{t,m}^i)}$ is the probability ratio, $s_{t,m}^i$ represents the concatenation of current state and the first $m-1$ tokens in the action sequence for agent $i$ at time step $t$, and $a_{t,m}^i$ denotes its $m$-th token. We compute the advantage estimate using GAE[15]: $\hat{A}_{t,m}^i = \sum_{l=0}^{M-m-1}(\gamma\lambda)^l \delta_{t,m+l}^i$, where $M$ is the token length of the action sequence.

To estimate state values across the multi-agent system, we employ a centralized state-value function $V_\phi$ that takes each agent's state $s_{t,m}^i$ as input. The value function is optimized to minimize the mean squared error:

$$\mathcal{L}_V^i(\phi) = \mathbb{E}_{\tau \in \pi_\theta}\left[\sum_t \sum_m (V_\phi(s_{t,m}^i) - \hat{G}_{t,m}^i)^2\right] \quad (7)$$

Where $\hat{G}_{t,m}^i = \hat{A}_{t,m}^i + V_\phi(s_{t,m}^i)$ is the empirical return. The final optimization objective combines the policy and value losses:

$$\mathcal{L}_{PPO}^i(\theta, \phi) = \mathcal{L}_{CLIP}^i(\theta) + c_v \mathcal{L}_V^i(\phi) \quad (8)$$

Where $c_v$ controls the weight of the value loss. This joint objective enables end-to-end training of both policy and value networks across all agents.

III. RESULTS&DISCUSSION

In this section, we first present the datasets and implementation details, followed by a comparison of our SIRAG with the latest RAG methods with analysis.

*A. Datasets*

To comprehensively evaluate our SIRAG, we experiment on both single-hop datasets including Natural Questions (NQ) and PopQA, as well as multi-hop datasets including 2WikiMultiHopQA (2Wiki), and HotpotQA (HQA). For each dataset, we only use 100 randomly sampled questions instead of the full training set.

*B. Experimental Details*

Following Asai et al.[8], we construct our retrieval system using the 2018 Wikipedia dump as the knowledge source and use contriever-msmarco as our dense retriever. We utilize Qwen2.5-7B-Instruct as fixed LLM server, while Qwen2-0.5B is trained as candidate lightweight agent for efficient edge deployment. In the warm-up phase, we collect 4 solutions for each question with Qwen2.5-7B-Instruct. We use a learning rate of 4e-5, with 2 epochs and a batch size of 4. For the RL phase, we set learning rate of 5e-7 for policy model and 5e-6 for value model with a batch size of 2 and maximal depth of 5.

## C. Performance Comparison

We report the performance on both single-hop and multi-hop datasets in Table 1. First, our SIRAG consistently outperforms various baselines across different datasets, achieving superior average performance of 48.23% with lightweight agents of only 0.5B parameters. This demonstrates the effectiveness of our agent-centric alignment approach in bridging the gap between the retriever and the LLM. Second, compared to single-hop datasets, our method yields particularly notable gains in challenging multi-hop reasoning tasks. Specifically, SIRAG achieves significant improvements on multi-hop datasets (2Wiki +9.3%, HQA +9.2%), while maintaining strong performance on single-hop tasks (PopQA +1.8%). This significant performance gain suggests that, even without training original RAG system, our SIRAG effectively enhances the coordination between the retriever and LLM, which is particularly crucial for addressing complex multi-hop tasks. Third, although retriever fine-tuning method requires fewer tuned parameters, it does not overcome the limitations of standard RAG systems in handling complex cognitive and multi-hop reasoning tasks. Both LLM fine-tuning and intermediate module methods show promising results, but are constrained by either large tuning parameters (7B/72B) or inconsistent performance across different reasoning datasets. In contrast, our SIRAG achieves consistent improvements across almost all datasets with only 0.5B additional parameters, demonstrating both efficiency and effectiveness in enhancing RAG systems.

Table 1 Main results. Comparison of EM(%) on four datasets.

| method | Agent | tuned params | 2Wiki | HQA | NQ | PopQA | Average |
|---|---|---|---|---|---|---|---|
| direct | / | - | 31.6 | 35.4 | 43.3 | 24.3 | 33.65 |
| standard | / | - | 27.3 | 42.1 | 50.8 | 30.0 | 37.55 |
| Reranker[19] | 7B | 7B | 26.4 | 37.2 | 47.6 | 20.9 | 33.02 |
| Query-Writer[6] | 1.5B | 1.5B | 32.3 | 36.8 | **53.1** | 30.5 | 38.18 |
| selfRAG[8] | / | 7B | 35.9 | 45.5 | 50.4 | 31.5 | 40.83 |
| SIRAG(ours) | 0.5b | 0.5b | **45.2** | **54.7** | 51.7 | **33.3** | **46.23** |

## D. Ablation Study

To thoroughly evaluate the effectiveness of different components in our training process, we conduct comprehensive ablation studies across four in-domain datasets. Specifically, we examine the following variants: (1) "w/o LLM judge": A variant without the tree-structured rollout and LLM-as-a-judge credit assignment, meaning that we directly optimize each agent using the system-level reward (a single trajectory). (2) "w/o RL": The performance in the supervised warm-up phase.

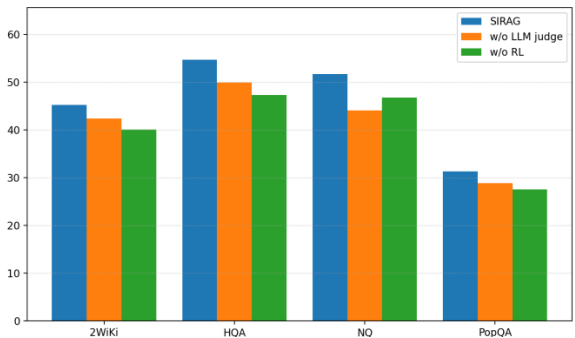

Figure 2 Ablation Study

The experimental results reveal several key findings. First, removing the tree-structured rollout (and LLM-as-a-judge credit assignment) leads to unstable performance during the RL phase, occasionally degrading below the supervised warm-up model. This degradation can be attributed to the direct use of system-level rewards as supervised signals for all agents, which fails to accurately assess individual agent contributions and may mask detrimental actions within successful trajectories. In contrast, our LLM-as-a-judge credit assignment mechanism enables reward allocation in probabilistic expectation through tree-structured exploration, ensuring that each agent receives appropriate feedback for its specific actions. Second, comparing with the supervised warm-up model ("w/o RL"), our SIRAG achieves substantial improvements across all datasets. This performance boost demonstrates that end-to-end RL optimization effectively aligns the behaviors of multiple agents towards the system-level objectives, going beyond the limitations of supervised learning that only optimizes for local agents.

## IV. CONCLUSIONS

In this paper, we proposed a process-supervised multi-agent framework for retrieval-augmented generation. By introducing two lightweight agents, the Decision Maker and the Knowledge Selector, our approach explicitly coordinates the interaction between the retriever and the generator. To address the challenge of sparse and unstable rewards, we employed an LLM-as-Judge to provide process-level supervision for each intermediate action, enabling more accurate credit assignment. Combined with a tree-structured rollout strategy and end-to-end optimization using PPO, our framework achieves stable training, interpretable reasoning trajectories, and improved accuracy on both single-hop and multi-hop QA tasks.

ACKNOWLEDGMENT

This work was supported by the International Science and Technology Cooperation Project of Guangzhou Economic and Technological Development District (No.2023GH16).

## REFERENCES

[1] Zhang, Y., Li, Y., Cui, L., Cai, D., Liu, L., Fu, T., Huang, X., Zhao, E., Zhang, Y., Chen, Y., et al. Siren's song in the ai ocean: a survey on hallucination in large language models. ArXiv abs/:2309.01219

[2] Sahoo, S. S., Plasek, J. M., Xu, H., Uzuner, Ö., Cohen, T., Yetisgen, M., Liu, H., Meystre, S., and Wang, Y. Large language models for biomedicine: foundations, opportunities, challenges, and best practices. Journal of the American Medical Informatics Association, pp. ocae074, 2024.

[3] Ji, Z., Lee, N., Frieske, R., Yu, T., Su, D., Xu, Y., Ishii, E., Bang, Y. J., Madotto, A., and Fung, P. Survey of hallucination in natural language generation. ACM Computing Surveys, 55(12):1–38, 2023.

[4] Lewis, P., Perez, E., Piktus, A., Petroni, F., Karpukhin, V., Goyal, N., Küttler, H., Lewis, M., Yih, W.-t., Rocktäschel, T., et al. Retrieval-augmented generation for knowledge intensive nlp tasks. Advances in Neural Information Processing Systems, 33:9459–9474, 2020.

[5] Gao, Y., Xiong, Y., Gao, X., Jia, K., Pan, J., Bi, Y., Dai, Y., Sun, J., and Wang, H. Retrieval-augmented generation for large language models: A survey. arXiv preprint arXiv:2312.10997, 2023.

[6] X. Ma, Y. Gong, P. He, H. Zhao, and N. Duan. Query rewriting in retrieval-augmented large language models. In H. Bouamor, J. Pino, and K. Bali, editors, Proceedings of the 2023 Conference on Empirical Methods in Natural Language Processing, EMNLP 2023, Singapore, December 6-10, 2023, pages 5303–5315. Association for Computational Linguistics, 2023. doi: 10.18653/V1/2023.EMNLP-MAIN.322. URL https://doi.org/10.18653/v1/2023.emnlp-main.322.

[7] W. Shi, S. Min, M. Yasunaga, M. Seo, R. James, M. Lewis, L. Zettlemoyer, and W. Yih. REPLUG: retrieval-augmented black-box language models. In K. Duh, H. Gómez-Adorno, and S. Bethard, editors, Proceedings of the 2024 Conference of the North American Chapter of the Association for Computational Linguistics: Human Language Technologies (Volume 1: Long Papers), NAACL 2024, Mexico City, Mexico, June 16-21, 2024, pages 8371–8384.

[8] A. Asai, Z. Wu, Y. Wang, A. Sil, and H. Hajishirzi. Self-rag: Learning to retrieve, generate, and critique through self-reflection. In The Twelfth International Conference on Learning Representations, ICLR 2024, Vienna, Austria, May 7-11, 2024. OpenReview.net, 2024.

[9] Z. Wei, W. Chen, and Y. Meng. Instructrag: Instructing retrieval-augmented generation with explicit denoising. CoRR, abs/2406.13629, 2024. doi: 10.48550/ARXIV.2406.13629.

[10] T. Yu, S. Zhang, and Y. Feng. Auto-rag: Autonomous retrieval-augmented generation for large language models. CoRR, abs/2411.19443, 2024. doi: 10.48550/ARXIV.2411.19443.

[11] Y. Yu, W. Ping, Z. Liu, B. Wang, J. You, C. Zhang, M. Shoeybi, and B. Catanzaro. Rankrag: Unifying context ranking with retrieval-augmented generation in llms. CoRR, abs/2407.02485, 2024. doi: 10.48550/ARXIV.2407.02485.

[12] E. Schmidt. How google works. Hachette UK, 2014.

[13] R. Nakano, J. Hilton, S. Balaji, J. Wu, L. Ouyang, C. Kim, C. Hesse, S. Jain, V. Kosaraju, W. Saunders, X. Jiang, K. Cobbe, T. Eloundou, G. Krueger, K. Button, M. Knight, B. Chess, and J. Schulman. Webgpt: Browser-assisted question-answering with human feedback. CoRR, abs/2112.09332, 2021.

[14] Y. Zhou, Y. Liu, X. Li, J. Jin, H. Qian, Z. Liu, C. Li, Z. Dou, T. Ho, and P. S. Yu. Trustworthiness in retrieval-augmented generation systems: A survey. CoRR, abs/2409.10102, 2024. doi: 10.48550/ARXIV.2409.10102.

[15] J. Schulman, P. Moritz, S. Levine, M. I. Jordan, and P. Abbeel. High-dimensional continuous control using generalized advantage estimation. In Y. Bengio and Y. LeCun, editors, 4th International Conference on Learning Representations, ICLR 2016, San Juan, Puerto Rico, May 2-4, 2016, Conference Track Proceedings, 2016.

[16] J. Schulman, F. Wolski, P. Dhariwal, A. Radford, and O. Klimov. Proximal policy optimization algorithms. CoRR, abs/1707.06347, 2017.

[17] L. Yuan, Z. Zhang, L. Li, C. Guan, and Y. Yu. A survey of progress on cooperative multi-agent reinforcement learning in open environment. CoRR, abs/2312.01058, 2023. doi: 10.48550/ ARXIV.2312.01058.

[18] C. Zhu, M. Dastani, and S. Wang. A survey of multi-agent deep reinforcement learning with communication. Auton. Agents Multi Agent Syst., 38(1):4, 2024. doi: 10.1007/S10458-023-09633-6.

[19] Z. Li, X. Zhang, Y. Zhang, D. Long, P. Xie, and M. Zhang. Towards general text embeddings with multi-stage contrastive learning. CoRR, abs/2308.03281, 2023. doi: 10.48550/ARXIV.

[20] A. Asai, Z. Wu, Y. Wang, A. Sil, and H. Hajishirzi. Self-rag: Learning to retrieve, generate, and critique through self-reflection. In The Twelfth International Conference on Learning Representations, ICLR 2024, Vienna, Austria, May 7-11, 2024. OpenReview.net, 2024.